\documentclass[10pt,twocolumn,letterpaper]{article}

\usepackage[pagenumbers]{cvpr} 

\usepackage{graphicx}
\usepackage{amsmath}
\usepackage{amssymb}
\usepackage{booktabs}
\usepackage{stmaryrd}
\usepackage{arydshln}

\newcommand{\modelname}{FlexEControl\xspace}

\usepackage{multirow}

\usepackage[pagebackref,breaklinks,colorlinks]{hyperref}

\usepackage[capitalize]{cleveref}
\crefname{section}{Sec.}{Secs.}
\Crefname{section}{Section}{Sections}
\Crefname{table}{Table}{Tables}
\crefname{table}{Tab.}{Tabs.}

\def\cvprPaperID{19} 
\def\confName{CVPR}
\def\confYear{2023}

\begin{document}

\title{FlexEControl: Flexible and Efficient Multimodal Control \\ for Text-to-Image Generation}

\author{%
 \textbf{ Xuehai He$^{1}$ \, Jian Zheng$^{2}$\,} \textbf{Jacob Zhiyuan Fang$^{2}$\, Robinson Piramuthu$^{2}$\,} \\ \textbf{Mohit Bansal$^{3}$\, Vicente Ordonez$^{4}$\, Gunnar A Sigurdsson$^{2}$\, Nanyun Peng$^{5}$\, Xin Eric Wang$^{1}$} \ \\ 
  $^1$UC Santa Cruz, $^2$Amazon, $^3$UNC Chapel Hill, $^4$Rice University, $^5$University of California, Los Angeles\\
  \texttt{\{xhe89,xwang366\}@ucsc.edu}\\}
\maketitle

\begin{abstract}
Controllable text-to-image (T2I) diffusion models generate images conditioned on both text prompts and semantic inputs of other modalities like edge maps.
Nevertheless, current controllable T2I methods commonly face challenges related to efficiency and faithfulness, especially when conditioning on multiple inputs from either the same or diverse modalities. 
In this paper, we propose a novel \emph{Flexible} and \emph{Efficient} method, \modelname, for controllable T2I generation.
At the core of \modelname is a unique weight decomposition strategy, which allows for streamlined integration of various input types. This approach not only enhances the faithfulness of the generated image to the control, but also significantly reduces the computational overhead typically associated with multimodal conditioning.
Our approach achieves a reduction of 41\% in trainable parameters and 30\% in memory usage compared with Uni-ControlNet. Moreover, it doubles data efficiency and can flexibly generate images under the guidance of multiple input conditions of various modalities. 
\end{abstract}

\section{Introduction}

\begin{figure}[!t]
     \centering
     \includegraphics[width=0.45\textwidth]{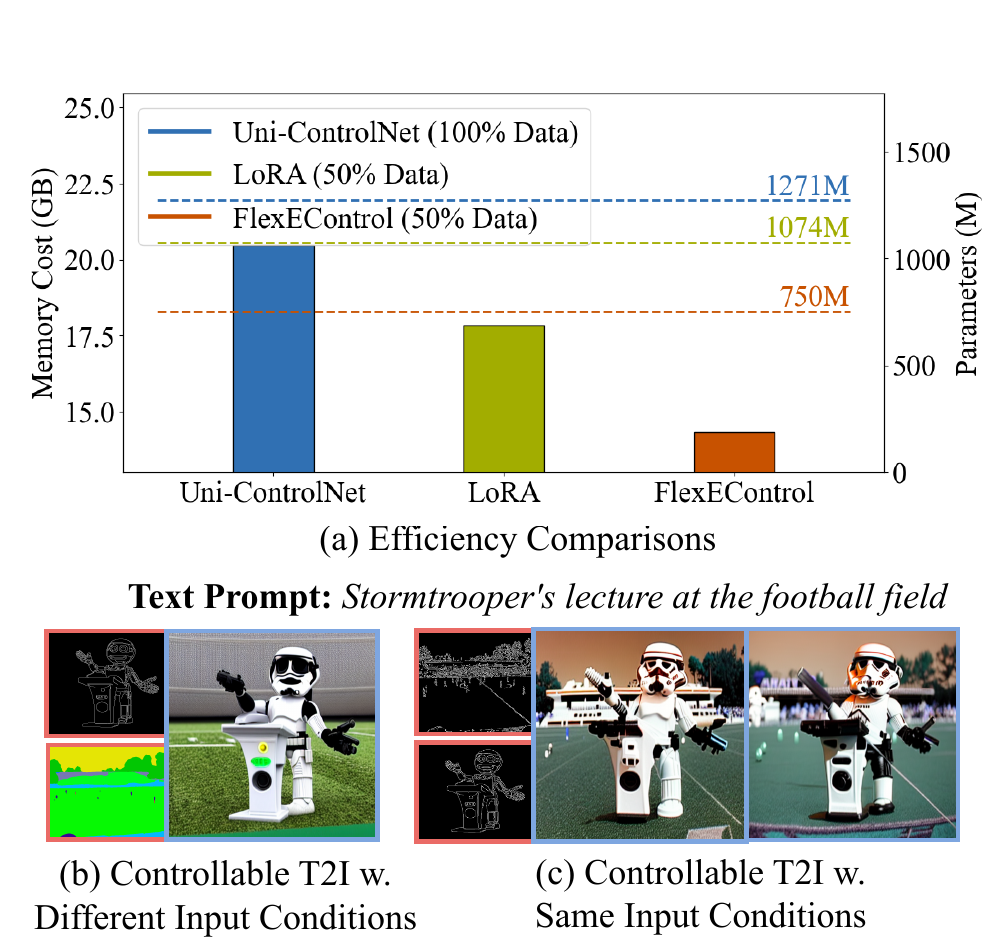}
     \caption{(a) \modelname excels in training efficiency, achieving superior performance with just half the training data compared to its counterparts on (b) Controllable Text-to-Image Generation w. Different Input Conditions (one edge map and one segmentation map). (c)  \modelname effectively conditions on two canny edge maps. The text prompt is \texttt{Stormtrooper's lecture at the football field} in both Figure (b) and Figure (c). }
     \label{fig:teaser}
\end{figure}

In the realm of text-to-image (T2I) generation, diffusion models exhibit exceptional performance in transforming textual descriptions into visually accurate images. Such models exhibit extraordinary potential across a plethora of applications, spanning from content creation~\cite{stable_diffusion,saharia2022photorealistic,nichol2021glide,ramesh2021zero,yu2022scaling,avrahami2023break,chang2023muse}, image editing~\cite{balaji2022ediffi,kawar2023imagic,couairon2022diffedit,zhang2023text,valevski2022unitune,nichol2021glide,hertz2022prompt,brooks2023instructpix2pix,mokady2023null}, and also fashion design~\cite{cao2023difffashion}. We propose a new unified method that can tackle two problems in text-to-image generation: improve the training efficiency of T2I models concerning memory usage, computational requirements, and a thirst for extensive datasets~\cite{imagen2022, stable_diffusion, dalle}; and improve their controllability especially when dealing with multimodal conditioning, e.g. multiple edge maps and at the same time follow the guidance of text prompts, as shown in~\autoref{fig:teaser} (c).  

Controllable text-to-image generation models~\cite{t2iadapter} often come at a significant training computational cost, with linear growth in cost and size when training with different conditions. Our approach can improve the training efficiency of existing text-to-image diffusion models and unify and flexibly handle different structural input conditions all together. We take cues from the efficient parameterization strategies prevalent in the NLP domain~\cite{pham2018efficient,huLoRALowRankAdaptation2021,zakenBitFitSimpleParameterefficient2021,houlsbyParameterEfficientTransferLearning2019} and computer vision literature~\cite{he2022parameter}.  The key idea is to learn shared decomposed weights for varied input conditions, ensuring their intrinsic characteristics are conserved.  Our method has several benefits: It not only achieves greater compactness~\cite{stable_diffusion}, but also retains the full representation capacity to handle various input conditions of various modalities; Sharing weights across different conditions contributes to the data efficiency; The streamlined parameter space aids in mitigating overfitting to singular conditions, thereby reinforcing the flexible control aspect of our model.

Meanwhile, generating images from multiple homogeneous conditional inputs, especially when they present conflicting conditions or need to align with specific text prompts, is challenging. To further augment our model’s capability to handle multiple inputs from either the same or diverse modalities as shown in Figure~\ref{fig:teaser}, during training, we introduce a new training strategy with two new loss functions introduced to strengthen the guidance of corresponding conditions. This approach, combined with our compact parameter optimization space, empowers the model to learn and manage multiple controls efficiently, even within the same category (e.g., handling two distinct segmentation maps and two separate edge maps).
Our primary contributions are summarized below: 
\begin{itemize}
    \item  We propose \modelname, a novel text-to-image generation model for efficient controllable image generation that substantially reduces training memory overhead and model parameters through decomposition of weights shared across different conditions.  
    \item We introduce a new training strategy to improve the flexible controllability of \modelname. Compared with previous works, \modelname can generate new images conditioning on multiple inputs from diverse compositions of multiple modalities.
    \item  \modelname  shows on-par performance with Uni-ControlNet~\cite{unicontrolnet} on controllable text-to-image generation with 41\% less trainable parameters and 30\% less training memory. Furthermore, \modelname exhibits enhanced data efficiency, effectively doubling the performance achieved with only half amount of training data.
\end{itemize}


\section{Method}
\begin{figure}[!t]
     \centering
     \includegraphics[width=0.45\textwidth]{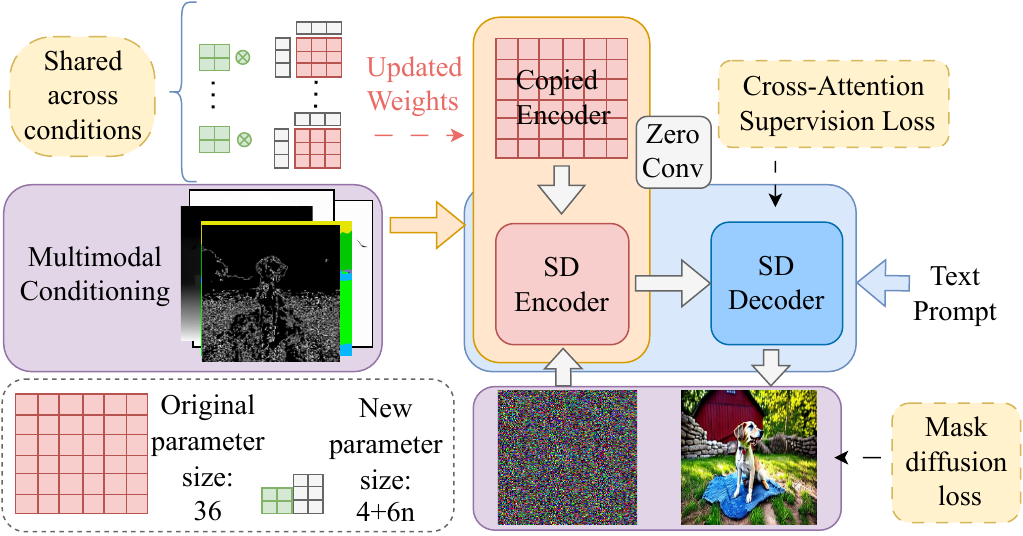}
     \caption{Overview of \modelname: a decomposed green matrix is shared across different input conditions, significantly enhancing the model's efficiency. During training, we integrate two specialized loss functions to enable flexible control and to adeptly manage conflicting conditions. In the example depicted here, the new parameter size is efficiently condensed to $4+6n$, where $n$ denotes the number of decomposed matrix pairs. }
     \label{fig:overview_method}
\end{figure}
The overview of our method is shown in Figure~\ref{fig:overview_method}. In general, we use the copied Stable Diffusion encoder which accepts structural conditional input and then perform efficient training via parameter reduction using Kronecker Decomposition first~\cite{phm} and then low-rank decomposition over the updated weights of the copied Stable Diffusion encoder. To enhance the control from language and different input conditions, we propose a new training strategy with two newly designed loss functions. The details are shown in the sequel.

\subsection{Preliminary}
We use Stable Diffusion 1.5~\cite{stable_diffusion} in our experiments. This model falls under the category of Latent Diffusion Models (LDM) that encode input images \( x \) into a latent representation \( z \) via an encoder \( \mathcal{E} \), such that \( z = \mathcal{E}(x) \), and subsequently carry out the denoising process within the latent space \( \mathcal{Z} \). An LDM is trained with a denoising objective as follows:

\begin{equation}
\mathcal{L}_{\text{ldm}} = \mathbb{E}_{z,c,e,t}\left[\left\lVert \hat{\epsilon}_\theta(z_t \mid c,t) - \epsilon \right\rVert^2\right]
\end{equation}
where $(z, c)$ constitute data-conditioning pairs (comprising image latents and text embeddings), $\epsilon \sim \mathcal{N}(0, I)$ , $t \sim \text{Uniform}(1, T)$, and $\theta$ denotes the model parameters.

\subsection{Efficient Training for Controllable Text-to-Image (T2I) Generation}
 Our approach is motivated by empirical evidence that Kronecker Decomposition~\cite{phm} effectively preserves critical weight information. We employ this technique to encapsulate the shared relational structures among different input conditions. Our hypothesis posits that by amalgamating diverse conditions with a common set of weights, data utilization can be optimized and training efficiency can be improved. We focus on decomposing and fine-tuning only the cross-attention weight matrices within the U-Net~\cite{unet} of the diffusion model, where recent works~\cite{custom_diffusion} show their dominance when customizing the diffusion model. As depicted in Figure~\ref{fig:overview_method}, the copied encoder from the Stable Diffusion will accept conditional input from different modalities. During training, we posit that these modalities, being transformations of the same underlying image, share common information. Consequently, we hypothesize that the updated weights of this copied encoder, $\Delta\boldsymbol{W}$, can be efficiently adapted within a shared decomposed low-rank subspace.  This leads to:
\begin{equation}
    \Delta\boldsymbol{W}=\sum_{i=1}^{n} \boldsymbol{H}_{\boldsymbol{i}} \otimes\left({u}_{{i}} {v}_{{i}}^{\top}\right)
\end{equation}
with $n$ is the number of decomposed matrices, ${u}_{{i}} \in \mathbb{R}^{\frac{k}{n} \times r}$ and ${v}_{{i}} \in \mathbb{R}^{r \times \frac{d}{n}}$, where $r$ is the rank of the matrix which is a small number, $\boldsymbol{H}_{\boldsymbol{i}}$ are the decomposed learnable matrices shared across different conditions, and $\otimes$ is the Kronecker product operation. The low-rank decomposition ensures a consistent low-rank representation strategy. This approach substantially saves trainable parameters, allowing efficient fine-tuning over the downstream text-to-image generation tasks.

\begin{figure}[!t]
     \centering
     \includegraphics[width=0.3\textwidth]{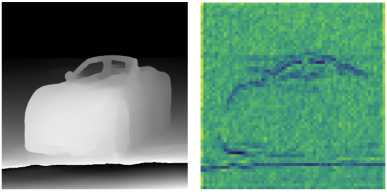}
     \caption{The visualization of decomposed shared “slow”
weights (right image) for single condition case where the input condition (left image) is the depth map and the input text prompt is \texttt{Car}. We took the average over the decomposed shared weights of the last cross-attention block across all attention heads in Stable Diffusion. }
     \label{map_visual}
\end{figure}

The intuition for why Kronecker decomposition works for finetuning partially is partly rooted in the findings of~\cite{phm,mahabadiCompacterEfficientLowRank2021,he2022parameter}. These studies highlight how the model weights can be broken down into a series of matrix products and thereby save parameter space. As shown in Figure~\ref{fig:overview_method}, the original weights is 6x6, then decomposed into a series of matrix products. When adapting the training approach based on the decomposition to controllable T2I, the key lies in the shared weights, which, while being common across various conditions, retain most semantic information. For instance, the shared “slow”
weights~\cite{fast_weights} of an image, combined with another set of  “fast” low-rank weights, can preserve the original image's distribution without a loss in semantic integrity, as illustrated in Figure \ref{map_visual}. This observation implies that updating the slow weights is crucial for adapting to diverse conditions. Following this insight, it becomes logical to learn a set of condition-shared decomposed weights in each layer, ensuring that these weights remain consistent across different scenarios. The data utilization and parameter efficiency is also improved.

\subsection{Enhanced Training for Conditional Inputs}
\label{sec:enhanced_training}
We then discuss how to improve the control under multiple input conditions of varying modalities with the efficient training approach.

\paragraph{Dataset Augmentation with Text Parsing and Segmentation}
To optimize the model for scenarios involving multiple homogeneous (same-type) conditional inputs, we initially augment our dataset. We utilize a large language model (\texttt{gpt-3.5-turbo}) to parse texts in prompts containing multiple object entities. The parsing query is structured as: \texttt{Given a sentence, analyze the objects in this sentence, give me the objects if there are multiple.} Following this, we apply CLIPSeg~\cite{clipseg} (\texttt{clipseg-rd64-refined} version) to segment corresponding regions in the images, allowing us to divide structural conditions into separate sub-feature maps tailored to the parsed objects.

\paragraph{Cross-Attention Supervision}
For each identified segment, we calculate a unified attention map, \(\boldsymbol{A}_i\), averaging attention across layers and relevant $N$ text tokens:
\begin{equation}
\boldsymbol{A}_i=\frac{1}{L}
\sum_{l=1}^L \sum_{i=1}^N \llbracket T_i \in \mathcal{T}_j \rrbracket \mathbf{CA}_i^l,
\end{equation}
where \(\llbracket \cdot \rrbracket\) is the Iverson bracket, \(\mathbf{CA}_i^l\) is the cross-attention map for token \(i\) in layer \(l\), and \(\mathcal{T}_j\) denotes the set of tokens associated with the \(j\)-th segment.

The model is trained to predict noise for image-text pairs concatenated based on the parsed and segmented results. An additional loss term, designed to ensure focused reconstruction in areas relevant to each text-derived concept, is introduced. Inspired by~\cite{breakascene}, this loss is calculated as the Mean Squared Error (MSE) deviation from predefined masks corresponding to the segmented regions:
\begin{equation}
\mathcal{L}_{\text{ca}} = \mathbb{E}_{z, t}\left[\left\|\boldsymbol{A}_i(v_i, z_t) - M_{i}\right\|_2^2\right],
\end{equation}
where \(\boldsymbol{A}_i(v_i, z_t)\) is the cross-attention map between token \(v_i\) and noisy latent \(z_t\), and \(M_{i}\) represents the mask for the \(i\)-th segment, which is derived from the segmented regions in our augmented dataset and appropriately resized to match the dimensions of the cross-attention maps.

\paragraph{Masked Noise Prediction}
To ensure fidelity to the specified conditions, we apply a condition-selective diffusion loss that concentrates the denoising effort on conceptually significant regions.  This focused loss function is applied solely to pixels within the regions delineated by the concept masks, which are derived from the non-zero features of the input structural conditions. Specifically, we set the masks to be binary where non-zero feature areas are assigned value of ones~\cite{multimodalgraphtransformer}, and areas lacking features are set to zero. Because of the sparsity of pose features for this condition, we use the all-ones mask. These masks serve to underscore the regions referenced in the corresponding text prompts:

\begin{equation}
\mathcal{L}_{\text{mask}} = \mathbb{E}_{z, \epsilon, t}\left[\left\|(\epsilon - \epsilon_\theta(z_t, t)) \odot M\right\|_2^2\right],
\end{equation}

where \(M\) represents the union of binary mask obtained from input conditions, $z_t$ denotes the noisy latent at timestep $t$, $\epsilon$ the injected noise, and $\epsilon_\theta$ the estimated noise from the denoising network (U-Net).

The total loss function employed is:
\begin{equation}
\mathcal{L}_{\text{total}} = \mathcal{L}_{\text{ldm}}  + \lambda_{\text{ca}} \mathcal{L}_{\text{ca}} + \lambda_{\text{mask}} \mathcal{L}_{\text{mask}},
\end{equation}
with $\lambda_{\text{rec}}$ and $\lambda_{\text{attn}}$ set to 0.01. The integration of $\mathcal{L}_{\text{ca}}$ and $\mathcal{L}_{\text{mask}}$ ensure the model will focus at reconstructing the conditional region and attend to guided regions during generation.

\section{Experiments}


\subsection{Datasets}~\label{dataset}
In pursuit of our objective of achieving controlled Text-to-Image (T2I) generation, we employed the LAION improved\_aesthetics\_6plus \cite{laion} dataset for our model training. Specifically, we meticulously curated a subset comprising 5,082,236 instances, undertaking the elimination of duplicates and applying filters based on criteria such as resolution and NSFW score. Given the targeted nature of our controlled generation tasks, the assembly of training data involved considerations of additional input conditions, specifically edge maps, sketch maps, depth maps, segmentation maps, and pose maps. The extraction of features from these maps adhered to the methodology expounded in\cite{controlnet}.



\subsection{Evaluation Metrics}
We employ a comprehensive benchmark suite of metrics including mIoU~\cite{miou}, SSIM~\cite{ssim}, mAP, MSE, FID~\cite{fid}, and CLIP Score~\cite{hessel2021clipscore,clip}~\footnote{https://github.com/jmhessel/clipscore}. The details are given in the Appendix.

\subsection{Experimental Setup}
\label{Experimental Setup}

In accordance with the configuration employed in Uni-ControlNet, we utilized Stable Diffusion 1.5~\footnote{https://huggingface.co/runwayml/stable-diffusion-v1-5} as the foundational model. Our model underwent training for a singular epoch, employing the AdamW optimizer~\cite{adam} with a learning rate set at $10^{-5}$. Throughout all experimental iterations, we standardized the dimensions of input and conditional images to $512\times512$. The fine-tuning process was executed on P3 AWS EC2 instances equipped with 64 NVIDIA V100 GPUs.

For quantitative assessment, a subset comprising 10,000 high-quality images from the LAION improved\_aesthetics\_6.5plus dataset was utilized. The resizing of input conditions to $512\times512$ was conducted during the inference process.

\subsubsection{Structural Input Condition Extraction}
We start from the processing of various local conditions used in our experiments. To facilitate a comprehensive evaluation, we have incorporated a diverse range of structural conditions, each processed using specialized techniques:

\begin{itemize}
\item Edge Maps: For generating edge maps, we utilized two distinct techniques:
\begin{itemize}
\item Canny Edge Detector~\cite{canny} - A widely used method for edge detection in images.
\item HED Boundary Extractor~\cite{hed} - Holistically-Nested Edge Detection, an advanced technique for identifying object boundaries.
\item MLSD~\cite{mlsd} - A method particularly designed for detecting multi-scale line segments in images.
\end{itemize}

 \item Sketch Maps: We adopted a sketch extraction technique detailed in~\cite{sketchmap} to convert images into their sketch representations.

 \item Pose Information: OpenPose~\cite{pose} was employed to extract human pose information from images, which provides detailed body joint and keypoint information.

 \item Depth Maps: For depth estimation, we integrated Midas~\cite{depth}, a robust method for predicting depth information from single images.

 \item Segmentation Maps: Segmentation of images was performed using the method outlined in~\cite{segmentation}, which focuses on accurately segmenting various objects within an image.
\end{itemize}




\subsection{Baselines}
In our comparative evaluation, we assess T2I-Adapter~\cite{t2iadapter}, PHM~\cite{phm}, Uni-ControlNet~\cite{unicontrolnet}, and LoRA~\cite{huLoRALowRankAdaptation2021}.

\begin{figure*}[t]
     \centering
     \includegraphics[width=\textwidth]{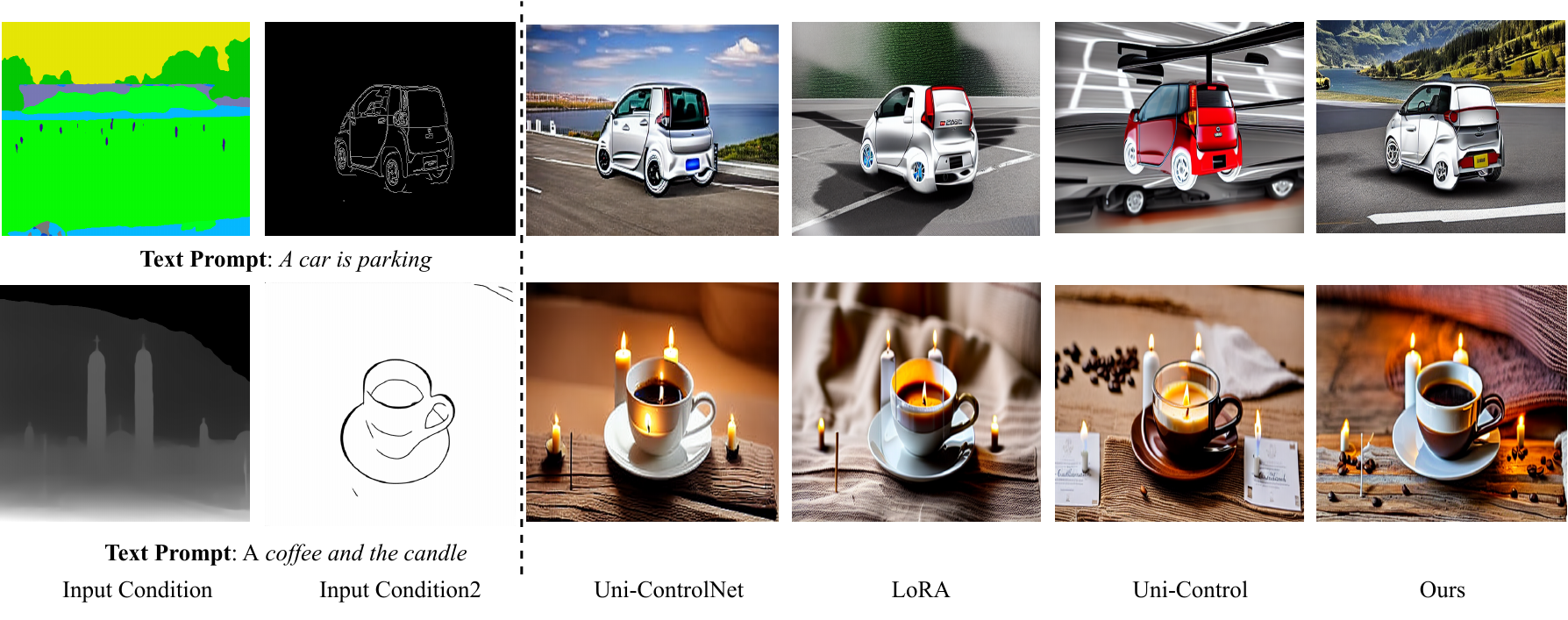}
     \caption{Qualitative comparison of \modelname and existing controllable diffusion models with multiple heterogeneous conditions. First row: \modelname effectively integrates both the segmentation and edge maps to generate a coherent image while Uni-ControlNet and LoRA miss the segmentation map and Uni-Control generates a messy image. Second row: The input condition types are one depth map and one sketch map. \modelname can do more faithful generation while all three others generate the candle in the coffee. }
     \label{fig:multi}
\end{figure*}

\begin{figure*}[!t]
     \centering
     \includegraphics[width=\textwidth]{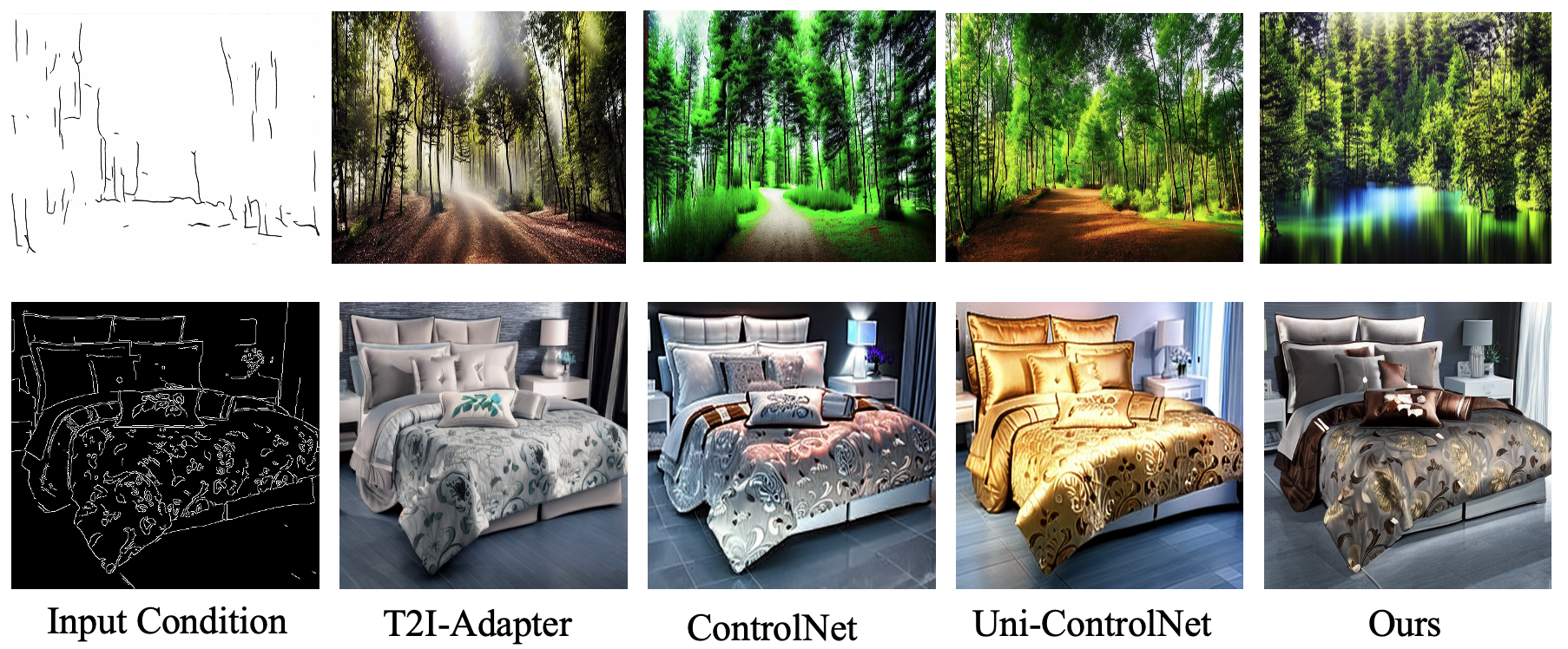}
     \caption{Qualitative comparison of \modelname and existing controllable diffusion models with single condition. Text prompt: \texttt{A bed.} The image quality of \modelname is comparable to existing methods and Uni-ControlNet + LoRA, while \modelname has much more efficiency.}
     \label{fig:single}
\end{figure*}

\begin{figure*}[!t]
     \centering
     \includegraphics[width=\textwidth]{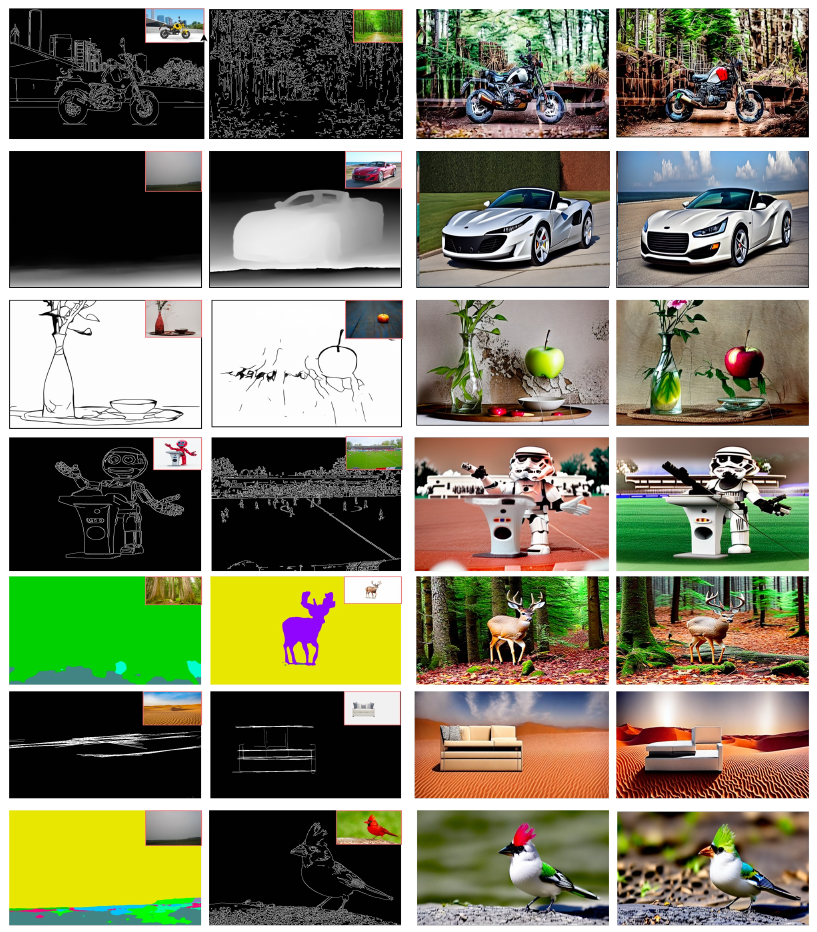}
     \caption{Qualitative performance of \modelname when conditioning on diverse compositions of multiple modalities. Each row in the figure corresponds to a unique type of condition, with the text prompts and conditions as follows: (first row) two canny edge maps with the prompt \texttt{A motorcycle in the forest}, (second row) two depth maps for \texttt{A car}, (third row) two sketch maps depicting \texttt{A vase with a green apple}, (fourth row) dual canny edge maps for \texttt{Stormtrooper's lecture at the football field}, (fifth row) two segmentation maps visualizing \texttt{A deer in the forests}, (sixth row) two MLSD edge maps for \texttt{A sofa in a desert}, and (seventh row) one segmentation map and one edge map for \texttt{A bird}. These examples illustrate the robust capability of \modelname to effectively utilize multiple multimodal conditions, generating images that are not only visually compelling but also faithfully aligned with the given textual descriptions and input conditions. }
     \label{fig:uniform}
\end{figure*}

\subsection{Quantitative Results}
\label{Results}

\begin{table}[!t]
\centering
\caption{Text-to-image generation efficiency comparison: \modelname shows substantial reductions in memory cost, trainable parameters, and training time, highlighting its improved training efficiency with the same model architecture. Training times are averaged over three runs up to 400 iterations for consistency.}

\label{tab:efficency-table}
\scalebox{0.75}{
\begin{tabular}{lcccc}
\toprule
Models & Memory Cost $\downarrow$ & \# Params. $\downarrow$ & Training Time $\downarrow$ \\

\midrule
  Uni-ControlNet~\cite{unicontrolnet} & 20.47GB & 1271M & 5.69 $\pm$ 1.33s/it \\
  LoRA~\cite{huLoRALowRankAdaptation2021} & 17.84GB & 1074M  & 3.97 $\pm$ 1.27 s/it \\
  PHM~\cite{phm} & 15.08GB & 819M  & 3.90 $\pm$ 2.01 s/it \\
\modelname (\textbf{ours}) & \textbf{14.33GB} & \textbf{750M} & \textbf{2.15 $\pm$ 1.42 s/it} \\

\bottomrule
\end{tabular}
}
\end{table}



\begin{table*}[ht]
\centering
\caption{Quantitative evaluation of controllability and image quality for single structural conditional inputs. \modelname performs overall better while maintaining much improved efficiency. }
\label{tab:table_performance}
\scalebox{0.9}{
\begin{tabular}{@{}lccccccccccc@{}}
\toprule
\multirow{2}{*}{Models} & Canny & MLSD & HED & Sketch & Depth & Segmentation & Poses & \multirow{2}{*}{FID$\downarrow$} & \multirow{2}{*}{CLIP Score$\uparrow$} \\
 & (SSIM)$\uparrow$ & (SSIM)$\uparrow$ & (SSIM)$\uparrow$ & (SSIM)$\uparrow$ & (MSE)$\downarrow$ & (mIoU)$\uparrow$ & (mAP)$\uparrow$ & \\
\midrule
T2IAdapter~\cite{t2iadapter} & 0.4480 & - &-&0.5241&90.01&0.6983&\textbf{0.3156}&27.80&0.4957\\
Uni-Control~\cite{unicontrol} &  0.4977& 0.6374&0.4885&0.5509&  90.04 & 0.7143 & 0.2083 &27.80 &0.4899\\
Uni-ControlNet~\cite{unicontrolnet} &  0.4910& 0.6083&0.4715&\textbf{0.5901}&  90.17 & 0.7084 & 0.2125 &27.74 &0.4890\\
PHM~\cite{phm} & 0.4365 & 0.5712 & 0.4633 & 0.4878  & 91.38 & 0.5534 & 0.1664 &27.91 &0.4961\\
LoRA~\cite{huLoRALowRankAdaptation2021} & 0.4497 & 0.6381 & \textbf{0.5043} & 0.5097  & \textbf{89.09} & 0.5480 & 0.1538 &27.99 &0.4832\\
\modelname(\textbf{ours}) &\textbf{0.4990} &\textbf{0.6385} &0.5041 &0.5518 & 90.93 &\textbf{0.7496}&0.2093 &\textbf{27.55}&\textbf{0.4963}\\
\bottomrule
\end{tabular}}
\end{table*}

\begin{table*}[ht]
\centering
\caption{Quantitative evaluation of controllability and image quality on~\modelname along with its variants and Uni-ControlNet. For Uni-ControlNet, we implement multiple conditioning by adding two homogeneous conditional images after passing through feature extractors.}
\label{tab:small_data}
\scalebox{0.75}{
\begin{tabular}{@{}lcccccccccccc@{}}
\toprule
& \multirow{2}{*}{Models} & Canny & MLSD & HED & Sketch & Depth & Segmentation & Poses & \multirow{2}{*}{FID$\downarrow$} & \multirow{2}{*}{CLIP Score$\uparrow$} \\
&& (SSIM)$\uparrow$ & (SSIM)$\uparrow$ & (SSIM)$\uparrow$ & (SSIM)$\uparrow$ & (MSE)$\downarrow$ & (mIoU)$\uparrow$ & (mAP)$\uparrow$ & & \\
\midrule
\multirow{4}{*}{Single Conditioning} & Uni-ControlNet & 0.3268&0.4097&0.3177&0.4096&98.80&0.4075&\textbf{0.1433}&29.43 &0.4844  \\
& \modelname (w/o $L_{ca}$)&0.3698 &0.4905 &0.3870 &0.4855 & 94.90 &0.4449&0.1432 &28.03 &0.4874\\
& \modelname (w/o $L_{mask}$) & 0.3701 & 0.4894 &0.3805&\textbf{0.4879}&\textbf{94.30} & 0.4418 &0.1432&28.19&0.4570\\
& \modelname &\textbf{0.3711} &\textbf{0.4920} &\textbf{0.3871} &0.4869 & 94.83 &\textbf{0.4479}&0.1432 &\textbf{28.03} &\textbf{0.4877}\\
\hdashline
\multirow{4}{*}{Multiple Conditioning} 
& Uni-ControlNet & 0.3078 & 0.3962 &0.3054&0.3871&98.84 & 0.3981 &0.1393&28.75&0.4828\\
& \modelname (w/o $L_{ca}$) & 0.3642 & 0.4901 &0.3704&0.4815&94.95 & 0.4368 &0.1405&28.50&0.4870\\
& \modelname (w/o $L_{mask}$) & 0.3666 & 0.4834 &0.3712&0.4831&94.89 & 0.4400 &0.1406&28.68&0.4542\\
& \modelname &\textbf{0.3690} &\textbf{0.4915} &\textbf{0.3784} &\textbf{0.4849} & \textbf{92.90} &\textbf{0.4429}&\textbf{0.1411} &\textbf{28.24} &\textbf{0.4873}\\
\bottomrule
\end{tabular}}
\end{table*}

Table~\ref{tab:efficency-table} highlights \modelname's superior efficiency compared to Uni-ControlNet. It achieves a 30\% reduction in memory cost, lowers trainable parameters by 41\% (from 1271M to 750M), and significantly reduces training time per iteration from 5.69s to 2.15s.


Table~\ref{tab:table_performance} provides a comprehensive comparison of \modelname's performance against Uni-ControlNet and T2IAdapter across diverse input conditions.  After training on a dataset of 5M text-image pairs, \modelname demonstrates better, if not superior, performance metrics compared to Uni-ControlNet and T2IAdapter. Note that Uni-ControlNet is trained on a much larger dataset (10M text-image pairs from the LAION dataset). Although there is a marginal decrease in SSIM scores for sketch maps and mAP scores for poses, \modelname excels in other metrics, notably surpassing Uni-ControlNet and T2IAdapter. This underscores our method's proficiency in enhancing efficiency and elevating overall quality and accuracy in controllable text-to-image generation tasks.

To substantiate the efficacy of \modelname in enhancing training efficiency while upholding commendable model performance, and to ensure a fair comparison, an ablation study was conducted by training models on an identical dataset. We traine~\modelname along its variants and Uni-ControlNet on a subset of 100,000 training samples from LAION improved\_aesthetics\_6plus. When trained with the identical data, \modelname performs better than Uni-ControlNet. The outcomes are presented in Table~\ref{tab:small_data}. Evidently, \modelname exhibits substantial improvements over Uni-ControlNet when trained on the same dataset. This underscores the effectiveness of our approach in optimizing data utilization, concurrently diminishing computational costs, and enhancing efficiency in the text-to-image generation process.

To validate \modelname's effectiveness in handling multiple structural conditions, we compared it with Uni-ControlNet through human evaluations. Two scenarios were considered: multiple homogeneous input conditions (300 images, each generated with 2 canny edge maps) and multiple heterogeneous input conditions (500 images, each generated with 2 randomly selected conditions). Results, summarized in Table~\ref{tab:humaneval}, reveal that \modelname was preferred by 64.00\% of annotators, significantly outperforming Uni-ControlNet (23.67\%). This underscores \modelname's proficiency with complex, homogeneous inputs. Additionally, \modelname demonstrated superior alignment with input conditions (67.33\%) compared to Uni-ControlNet (23.00\%). In scenarios with random heterogeneous conditions, \modelname was preferred for overall quality and alignment over Uni-ControlNet.

In addition to our primary comparisons, we conducted an additional quantitative evaluation of~\modelname and Uni-ControlNet. This evaluation focused on assessing image quality under scenarios involving multiple conditions from both the homogeneous and heterogeneous modalities. The findings of this evaluation are summarized in Table~\ref{tab:quantitative_results_diverse_modalities}. \modelname consistently outperforms Uni-ControlNet in both categories, demonstrating lower FID scores for better image quality and higher CLIP scores for improved alignment with text prompts.

\begin{table}[!t]
\centering
\caption{Human evaluation of \modelname and Uni-ControlNet under homogenous and heterogeneous structural conditions, assessing both human preference and condition alignment. "Win" indicates \modelname's preference, "Tie" denotes equivalence, and "Lose" indicates Uni-ControlNet's preference. Results indicate that under homogeneous conditions, \modelname outperforms Uni-ControlNet in both human preference and condition alignment.}
\label{tab:humaneval}
\medskip
\scalebox{0.8}{
\begin{tabular}{lccccc}
\toprule
Condition Type & Metric & Win & Tie & Lose \\
\midrule
\multirow{2}{*}{Homogeneous} & Human Preference (\%)& \textbf{64.00} & 12.33 & 23.67 \\
& Condition Alignment (\%)& \textbf{67.33} & 9.67 & 23.00 \\
\midrule
\multirow{2}{*}{Heterogeneous } & Human Preference (\%)&\textbf{9.80}&87.40&2.80 \\
& Condition Alignment (\%)&\textbf{6.60} & 89.49 &4.00 \\
\bottomrule
\end{tabular}}
\end{table}

\begin{table}[!t]
\centering
\caption{Quantitative evaluation of controllability and image quality in scenarios with multiple conditions from heterogeneous and homogeneous modalities for \modelname and Uni-ControlNet. The 'heterogeneous' category averages the performance across one Canny condition combined with six other different modalities. The 'homogeneous' category represents the average performance across seven identical modalities (three inputs). }
\label{tab:quantitative_results_diverse_modalities}
\medskip
\scalebox{0.95}{
\begin{tabular}{lcccc}
\toprule
Condition Type & Baseline & FID$\downarrow$ & CLIP Score$\uparrow$ \\
\midrule
\multirow{2}{*}{Heterogeneous} & Uni-ControlNet  & 27.81 & 0.4869 \\
& FlexEControl& \textbf{27.47} & \textbf{0.4981} \\
\hdashline
\multirow{2}{*}{Homogeneous} & Uni-ControlNet & 28.98 & 0.4858 \\
& FlexEControl& \textbf{27.65} & \textbf{0.4932} \\
\bottomrule
\end{tabular}}
\end{table}

\subsection{Qualitative Results}
We present qualitative results of our \modelname under three different settings: single input condition, multiple heterogeneous conditions, and multiple homogeneous conditions, illustrated in Figure~\ref{fig:single}, Figure~\ref{fig:multi}, and Figure~\ref{fig:uniform}, respectively. The results indicate that \modelname is comparable to baseline models when a single condition is input. However, with multiple conditions, \modelname consistently and noticeably outperforms other models. Particularly, under multiple homogeneous conditions, \modelname excels in generating overall higher quality images that align more closely with the input conditions, surpassing other models.




\section{Related Work}
\modelname is an instance of efficient training and controllable text-to-image generation. Here, we overview modeling efforts in the subset of efficient training towards reducing parameters and memory cost and controllable T2I.

\paragraph{Efficient Training} Prior work has proposed efficient training methodologies both for pretraining and fine-tuning. These methods have established their efficacy across an array of language and vision tasks. One of these explored strategies is Prompt Tuning~\cite{prompt_tuning}, where trainable prompt tokens are appended to pretrained models~\cite{first_prompt, promptingvl, visualprompttuning,he2022cpl}. These tokens can be added exclusively to input embeddings or to all intermediate layers~\cite{liPrefixTuningOptimizingContinuous2021}, allowing for nuanced model control and performance optimization. Low-Rank Adaptation (LoRA)~\cite{huLoRALowRankAdaptation2021} is another innovative approach that introduces trainable rank decomposition matrices for the parameters of each layer. LoRA has exhibited promising fine-tuning ability on large generative models including diffusion models~\cite{discffusion}, indicating its potential for broader application. Furthermore, the use of Adapters inserts lightweight adaptation modules into each layer of a pretrained transformer~\cite{houlsbyParameterEfficientTransferLearning2019, ruckleAdapterDropEfficiencyAdapters2021}. This method has been successfully extended across various setups~\cite{zhangTipAdapterTrainingfreeCLIPAdapter2021,clip_adapter,t2iadapter}, demonstrating its adaptability and practicality. Other approaches including post-training model compression~\cite{structural_prunning} facilitate the transition from a fully optimized model to a compressed version -- either sparse~\cite{sparsegpt}, quantized~\cite{Q-diffusion,vector_quantized_diffusion}, or both. This methodology was particularly helpful for parameter quantization~\cite{dettmers2023qlora}. Different from these methodologies, our work puts forth a new unified strategy that aims to enhance the efficient training of text-to-image diffusion models through the leverage of low-rank structure. Our proposed method integrates principles from these established techniques to offer a fresh perspective on training efficiency, adding to the rich tapestry of existing solutions in this rapidly evolving field.

\paragraph{Controllable Text-to-Image Generation}
Recent developments in the text-to-image generation domain strives for more control over image generation, enabling more targeted, stable, and accurate visual outputs, several models like T2I-Adapter~\cite{t2iadapter} and Composer~\cite{composer} have emerged to enhance image generations following the semantic guidance of text prompts and multiple different structural conditional control. However, existing methods are struggling at dealing with multiple conditions from the same modalities, especially when they have conflicts, e.g. multiple segmentation maps and at the same time follow the guidance of text prompts; Recent studies also highlight challenges in controllable text-to-image generation (T2I), such as omission of objects in text prompts and mismatched attributes \cite{lee2023aligning, Bakr_2023_ICCV}, showing that current models are strugging at handling controls from different conditions. Towards these, the Attend-and-Excite method \cite{Chefer2023AttendandExciteAS} refines attention regions to ensure distinct attention across separate image regions. ReCo \cite{yang2022reco}, GLIGEN \cite{Li2023GLIGENOG}, and Layout-Guidance \cite{chen2023trainingfree} allow for image generation informed by bounding boxes and regional descriptions. Our work improves the model's controllability by proposing a new training strategy.

\section{Conclusion}
This work introduces a unified approach that improves both the flexibility and efficiency of diffusion-based text-to-image generation. Our experimental results demonstrate a substantial reduction in memory cost and trainable parameters without compromising inference time or performance. 
Future work may explore more sophisticated decomposition techniques, furthering the pursuit of an optimal balance between model efficiency, complexity, and expressive power.

\clearpage
{\small
\bibliographystyle{ieee_fullname}
\bibliography{main}
}

\end{document}


\title{FlexEControl: Flexible and Efficient Multimodal Control \\ for Text-to-Image Generation}

\author{First Author\\
Institution1\\
Institution1 address\\
{\tt\small firstauthor@i1.org}
\and
Second Author\\
Institution2\\
First line of institution2 address\\
{\tt\small secondauthor@i2.org}
}
\maketitle

\section{Additional Implementation Details}
\label{architecture}
In this section, we provide further details about the implementation aspects of our approach.

In ControlNet~\cite{controlnet} and Uni-ControlNet~\cite{unicontrolnet}, the weights of Stable Diffusion (SD)~\cite{stable_diffusion} are fixed and the input conditions are fed into zero-convolutions and added back into the main Stable Diffusion backbone. Specifically, for Uni-ControlNet, they uses a multi-scale condition injection strategy that extracts features at different resolutions and uses them for condition injection referring to the implementation of Feature Denormalization (FDN):
\begin{align}
&\text{FDN}\left(Z, c\right) = \operatorname{norm}\left(Z\right) \cdot \left(1+\Phi\left(\operatorname{zero}\left(h_r\left(c\right)\right)\right)\right) \nonumber \\
&+ \Phi\left(\operatorname{zero}\left(h_r\left(c\right)\right)\right),
\end{align}
where $Z$ denotes noise features, $c$ denotes the input conditional features, $\Phi$ denotes learnable convolutional layers, and $\operatorname{zero}$ denotes zero convolutional layer.  The zero convolutional layer contains weights initialized to zero. This ensures that during the initial stages of training, the model relies more on the knowledge from the backbone part, gradually adjusting these weights as training progresses. The use of such layers aids in preserving the architecture's original behavior while introducing structure-conditioned inputs. We use the similar model architecture while we perform efficient training proposed in the main paper. We show the model architecture in Figure~\ref{fig:architecture}.
\begin{figure}[!t]
     \centering
     \includegraphics[width=0.3\textwidth]{NAACL2024/model_architecture.pdf}
     \caption{Detailed model architecture in \modelname. The Stable Diffusion part is fixed and others are trainable.}
     \label{fig:architecture}
\end{figure}

\subsection{Structural Input Condition Extraction}
We start from the processing of various local conditions used in our experiments. To facilitate a comprehensive evaluation, we have incorporated a diverse range of structural conditions, each processed using specialized techniques:

\begin{itemize}
    \item \textbf{Edge Maps}: For generating edge maps, we utilized two distinct techniques:
        \begin{itemize}
            \item Canny Edge Detector~\cite{canny} - A widely used method for edge detection in images.
            \item HED Boundary Extractor~\cite{hed} - Holistically-Nested Edge Detection, an advanced technique for identifying object boundaries.
            \item MLSD~\cite{mlsd} - A method particularly designed for detecting multi-scale line segments in images.
        \end{itemize}

    \item \textbf{Sketch Maps}: We adopted a sketch extraction technique detailed in~\citet{sketchmap} to convert images into their sketch representations.

    \item \textbf{Pose Information}: OpenPose~\cite{pose} was employed to extract human pose information from images, which provides detailed body joint and keypoint information.

    \item \textbf{Depth Maps}: For depth estimation, we integrated Midas~\cite{depth}, a robust method for predicting depth information from single images.

    \item \textbf{Segmentation Maps}: Segmentation of images was performed using the method outlined in~\citet{segmentation}, which focuses on accurately segmenting various objects within an image.
\end{itemize}

Each of these conditions plays a crucial role in guiding the text-to-image generation process, helping \modelname to generate images that are not only visually appealing but also semantically aligned with the given text prompts and structural conditions.

\section{Baselines}
For baseline implementations, we compare~\modelname with T2I-Adapter~\cite{t2iadapter}, PHM~\cite{phm}, Uni-ControlNet~\cite{unicontrolnet}, and LoRA~\cite{huLoRALowRankAdaptation2021} where we implement LoRA and PHM layers over the trainable modules in Uni-ControlNet interms of generated image quality and controllability. The rank of LoRA is set to 4. For PHM~\cite{phm}, we implement it by performing Kronecker decomposition and share weights across different layer, with the number of decomposed matrix being 4.

\section{Evaluation Metrics}
\noindent\textbf{mIoU~\cite{miou}:} Mean Intersection over Union, a metric that quantifies the degree of overlap between predicted and actual segmentation maps.
    
    \noindent\textbf{SSIM~\cite{ssim}:} Structural Similarity, a metric evaluating the structural similarity in generated outputs, applied to Canny edges, HED edges, MLSD edges, and sketches.
    
    \noindent\textbf{mAP:} Mean Average Precision, utilized for pose maps, measuring the precision of localization across multiple instances.
    
    \noindent\textbf{MSE:} Mean Squared Error, employed for depth maps, MSE quantifies the pixel-wise variance, providing an assessment of image fidelity.
    
    \noindent\textbf{FID~\cite{fid}:} Fréchet Inception Distance, which serves as a metric to quantify the realism and diversity of the generated images. A lower FID value indicates higher quality and diversity of the output images.
    
    \noindent\textbf{CLIP Score~\cite{hessel2021clipscore,clip}~\footnote{https://github.com/jmhessel/clipscore}:} Employing CLIP Score, we gauge the semantic similarity between the generated images and the input text prompts.

\clearpage
{\small
\bibliographystyle{ieee_fullname}
\bibliography{egbib}
}